\documentclass[letterpaper, 10 pt, conference]{ieeeconf}  
\IEEEoverridecommandlockouts                            % This command is only needed if 
\let\labelindent\relax
\usepackage{amsfonts}       % An extended set of fonts for use in mathematics

\usepackage{amsthm}
\usepackage{mathtools}      % Loads amsmath
\usepackage{amssymb}        % provides AMS fonts and symbols ex. \mathbb{...}
\usepackage{filecontents}
\usepackage{graphicx}       % enhanced support for graphics
\usepackage{marginnote}     % Add notes
\usepackage{marvosym}       % mvat
\usepackage{overpic}        % Add text on figures
\usepackage{tabularx}
\usepackage{subcaption}
\usepackage{cite}
\usepackage{color}
\usepackage[normalem]{ulem}
\usepackage{epstopdf}
\usepackage{enumitem}
\usepackage{lipsum}
\usepackage{url}
\usepackage{hyperref}
\usepackage[dvipsnames]{xcolor}
\usepackage[font=small,labelfont=bf]{caption}
\usepackage{diagbox}
\usepackage{multirow}
\usepackage{xspace}

\makeatletter
\newif\if@restonecol
\makeatother

\usepackage[linesnumbered,ruled,noend]{algorithm2e}%[ruled,vlined]{
\usepackage{algpseudocode}
\usepackage{amsmath}

\newif\iftwocolumn
\twocolumntrue

\newtheorem{theorem}{Theorem}[section]
\newtheorem*{remark}{Remark}

\makeatletter
\def\subsubsection{\@startsection{subsubsection}% name
                                 {3}% level
                                 {\z@ \hspace*{1mm}}% indent (formerly \parindent)
                                 {0ex plus 0.1ex minus 0.1ex}% before skip
                                 {0ex}% after skip
                                 {\normalfont\normalsize\itshape}}% style
\makeatother

\newcommand{\mpp}{\textsc{MRPP}\xspace}

\newcommand{\ecbs}{\textsc{ECBS}\xspace}
\newcommand{\ecbsd}{\textsc{ECBS}d\xspace}
\newcommand{\ddm}{\textsc{DDM}\xspace}
\newcommand{\dmpp}{\textsc{MRPS}\xspace}
\newcommand{\ryp}{\textsc{PRYP}\xspace}  %or anny other cool name
\newcommand{\eryp}{\textsc{EPRY}\xspace}
\newcommand{\orca}{\textsc{ORCA}\xspace}

\font\titlefont=ptmb at 14.7pt
\title{\titlefont Bin Assignment and Decentralized Path Planning for Multi-Robot Parcel Sorting
}

\author{Teng Guo \qquad Si Wei Feng   \qquad Jingjin Yu% <-this % stops a space
\thanks{G. Teng, S. Feng and J. Yu are with the Department of 
Computer Science, Rutgers, the State University of New Jersey, Piscataway, NJ, USA. 
Emails: {\tt\small \{ teng.guo, siwei.feng, jingjin.yu\}@rutgers.edu}.
}
}

\begin{document}

\maketitle
\thispagestyle{empty}
\pagestyle{empty}

\begin{abstract}
At modern warehouses, mobile robots transport packages and drop them into collection bins/chutes based on shipping destinations grouped by, e.g., the ZIP code. 
System throughput, measured as the number of packages sorted per unit of time, determines the efficiency of the warehouse. 
% 
%The throughput is largely determined by the average traveling cost to deliver a package. 
% 
This research develops a scalable, high-throughput multi-robot parcel sorting (\dmpp) solution, decomposing the task into two related processes, bin assignment and offline/online multi-robot path planning, and optimizing both.
Bin assignment matches collection bins with package types to minimize traveling costs. Subsequently, robots are assigned to pick up and drop packages to assigned bins. 
%While \dmpp is the online version of multi-robot path planning where robots are immediately assigned a new goal upon reaching their current one. 
% 
%Due to the NP-hardness  of multi-robot path planning, it is of key importance to address the trade-off  between solution quality and scalability for large-scale, real-world applications.
% 
Multiple highly effective bin assignment algorithms are proposed that can work with an arbitrary planning algorithm.
We propose a decentralized path planning routine using only local information to route the robots over a carefully constructed directed road network for multi-robot path planning. 
%
% Our decentralized planner achieves near-optimal results comparable to best centralized planners while dramatically reducing computation times by orders of magnitude.
Our decentralized planner, provably probabilistically deadlock-free, consistently delivers near-optimal results on par with some top-performing centralized planners while significantly reducing computation times by orders of magnitude.
%achieves a good balance between solution quality and scalability as compared with centralized algorithms for \dmpp.
% 
%In addition, unlike centralized planners which require the global information of the environment, \ryp can resolve conflicts using local information.
Extensive simulations show that our overall framework delivers promising performances.
%\noindent Simulation video: \url{https://youtu.be/B3pa4R4s93A} \textcolor{red}{The video needs to be updated to reflect ICRA}\\
\noindent Upon the publication of the manuscript, source code and data will be released at \url{https://github.com/arc-l/mrps}
\end{abstract}

\section{Introduction}
\begin{figure}[h]
\vspace{2mm}
    \centering
\begin{overpic}[width=0.98\columnwidth,tics=5]
{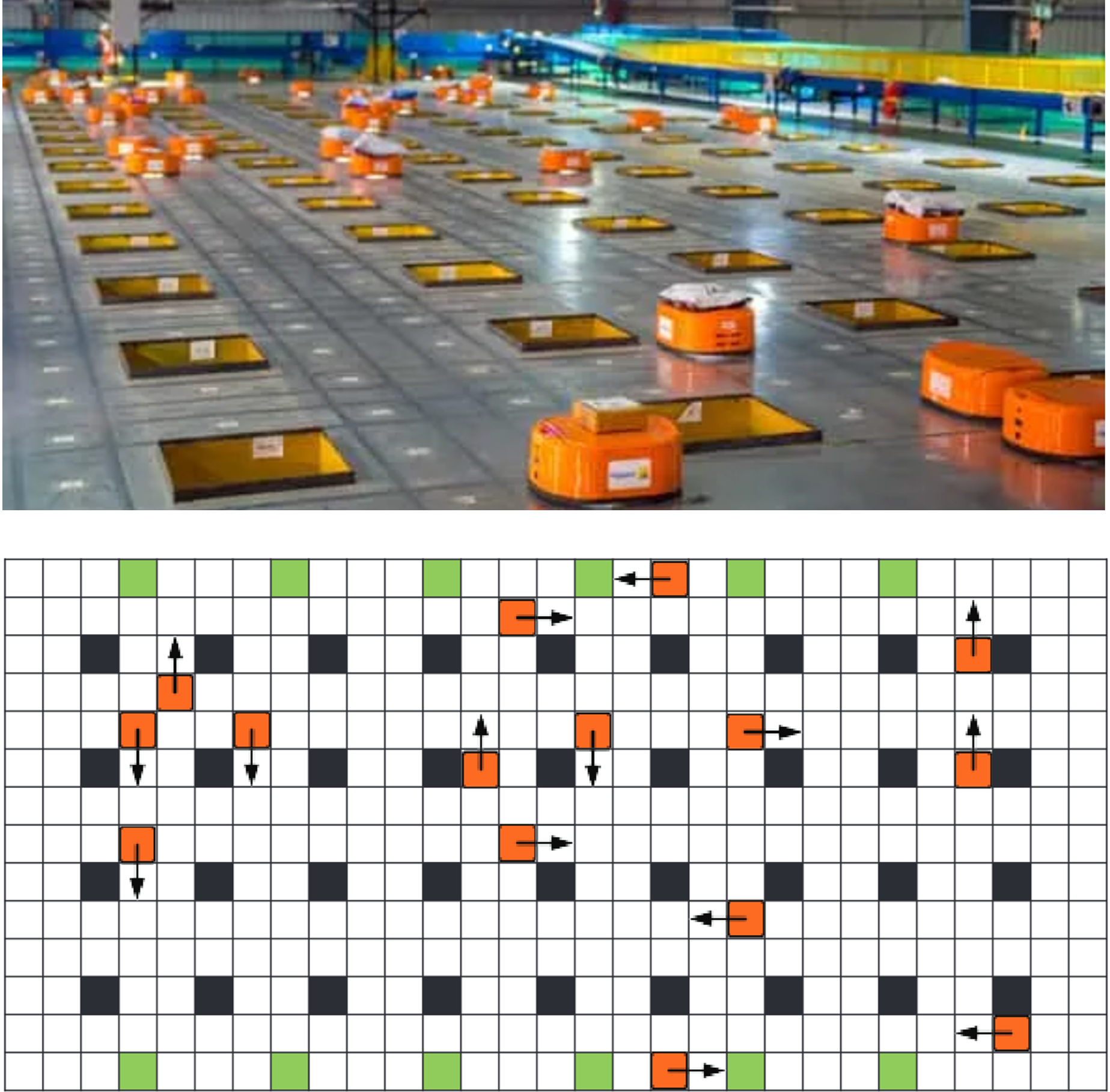}
%\put(15, -6){{\small (a)}}
%\put(70, -6){{\small (b)}}
\end{overpic}
\vspace{1mm}
    \caption{[top] Many mobile robots moving packages in a sortation center. In this setup, orange mobile robots are loaded with parcels and must drop them into specific chutes. Sortation centers like this are used by leading online merchants across the globe. [bottom] The graph abstraction of a sortation center. Black cells represent parcel collection bins and green cells package pickup stations. }
    \label{fig:sorting_center}
    \vspace{-1mm}
\end{figure}

% 
%In today's e-commerce systems, a sortation center is used to sort items in warehouses or distribution centers based on their destinations.
% 
%Fig.~\ref{fig:sorting_center}(b) shows the layout of a typical warehouse sortation center.
%
At autonomous sortation centers, a layout of which is illustrated in Fig.~\ref{fig:sorting_center}, robots move to assigned stations to pick up packages and then to the destination sorting bins/chutes to drop these packages.
A given bin usually collects parcels going to shipping addresses grouped, e.g., by the ZIP code.
%After arriving at the destination, the robot drops the parcel into the bin.
% 
After a robot successfully delivers a parcel, it is assigned another task. 
We denote this two-phase, dynamic planning problem as \emph{multi-robot parcel sorting} (\dmpp).

\emph{Throughput}, the average number of parcels delivered per timestep over a period of time, is a common criterion for evaluating the performance of \dmpp systems.
Assuming the average traveling cost for sorting one parcel is $\Bar{d}$, the throughput for $n$ robots equals $n/\Bar{d}$.
Therefore, reducing $\Bar{d}$ increases the throughput, which can be achieved through improving two sub-processes.
First, we strategically select the bins assigned to robots to reduce robot travel costs.
This helps especially when the distribution of each type of parcel is known to be imbalanced.
This forms the \emph{bin assignment problem}.
Following bin assignment, \emph{multi-robot path planning} (\mpp) must be carefully performed, which is NP-hard to optimize \cite{yu2013multi,surynek2012towards}.
Due to the challenge, computing optimal routing solutions for large-scale problems is generally impractical.
Instead, a fast sub-optimal planning algorithm with good optimality is generally preferred. 

%$\textcolor{red}{JJ's editing location}

%\textbf{Results and Contributions.}
Committing to the two-phase approach, we propose efficient methods for addressing each. Together, these methods result in an efficient algorithmic framework for \dmpp. The main results and contributions of our work are:
\begin{itemize}[leftmargin=3.3mm]
\item Algorithmically, we model the bin assignment task as an optimal assignment problem matching bins with specific parcel types. 
Optimal/sub-optimal algorithms, including a high-performance genetic algorithm, are proposed to solve the resulting assignment problem, resulting in much-improved system throughput when combined with an off-the-shelf multi-robot path planner. 
\vspace{1mm}
\item Leveraging the regularity of grid-like warehouses, a directed network is imposed that comes with multiple throughput-enhancing properties. 
An efficient \emph{decentralized} dynamic path planner running over the network directly prioritizes based on the inherent assignment order. The solution, provably probabilistically deadlock-free, greatly speeds up the planning process while simultaneously realizing high levels of solution optimality compared to SOTA centralized algorithms. 
\item
We benchmark on multiple popular SOTA centralized and decentralized algorithms, facilitating future theoretical and algorithmic studies of the \dmpp problem. 
\end{itemize}

\textbf{Ralated Work}. Multi-robot path planning (\mpp) has been extensively studied. In static/one-shot settings \cite{stern2019multi},  given a (graph) environment and many robots, with each robot having unique start and goal vertices,  the task is to find collision-free paths for routing all the robots.
Solving one-shot \mpp optimally in terms of minimizing either makespan or sum-of-cost is NP-complete \cite{surynek2009novel,yu2013structure}.
Subsequently, multi-robot parcel sorting (\dmpp), an online variant of \mpp where the robot would be assigned a new goal after reaching their current goal, is also NP-hard to optimally solve.\footnote{This can be readily proven by reducing viewing \mpp as a special type of \dmpp. I.e.,\mpp is a restricted \dmpp.}

\mpp solvers can be \emph{centralized} and \emph{decentralized}.  
Centralized solvers assume robots' paths can be calculated in a central computation node and subsequently executed without coordination error among the robots. Decentralized solvers assume each robot has significant autonomy and calculates the path independently, with necessary coordination among the robots to facilitate decision-making.  Centralized solvers either reduce \mpp to other well-studied problems \cite{yu2016optimal,surynek2010optimization,erdem2013general} or use search algorithms to search the joint space to find the solution \cite{sharon2015conflict,barer2014suboptimal,sharon2013increasing,silver2005cooperative}. Recently, polynomial time $1.x$-optimal algorithms potentially suitable for large-scale parcel sorting have also been proposed \cite{GuoYuRSS22,GuoFenYu22IROS}. 
Many researchers also looked into machine learning methods to directly learn decentralized policies for \mpp \cite{sartoretti2019primal,li2020graph}.

A common approach to solving \dmpp ``stitches'' one-shot \mpp instances together by using a (usually complete, bounded sub-optimal) \mpp planner to \emph{recompute} paths at each timestep at least one robot is assigned a new goal \cite{vcap2015complete}.
Replanning can be time-consuming as resources are wasted in redundant path computations. 
Han et. al\cite{han2020ddm} use a pre-computed database to resolve collisions scalably. However, as it only uses local information, the solution quality worsens in dense environments. 
Some planners plan new paths only for robots with new goals, which also lacks global oversight \cite{Ma2017LifelongMP}. Another promising method plans paths within a finite window, leading to better scalability \cite{li2020lifelong} at the cost of completeness. 
This phenomenon worsens when the planning window size is small compared to the average distance to the goal, as robots cannot predict the situation outside the planning window and might plan greedy short-term paths, resulting in unsolvable scenarios.

\textbf{Organization.}
In Section \ref{sec:preliminaries}, the environment setting and problem definition are given. 
Section \ref{sec:bin_assignment} outlines the three proposed bin assignment algorithms. Section \ref{sec:dmrpp} describes a near-optimal multi-robot planning algorithm, probabilistically deadlock-free, that can run in a decentralized manner. In Section \ref{sec:evaluation}, we provide extensive evaluation results of the proposed  \dmpp algorithms. We conclude in Section \ref{sec:conclusion}.
\section{Multi-Robot Parcel Sorting Problem}\label{sec:preliminaries}
Multi-robot parcel sorting (\dmpp) consists of \emph{bin assignment} and \emph{multi-robot path planning} in an intelligent sorting warehouse.
A warehouse is defined as a  grid world $\mathcal{G}(\mathcal{V},\mathcal{E})$ with $n$ mobile robots, $n_b$  bins, and $n_p$ pickup stations where robots may receive packages to be dropped off in bins. 
To make the problem more concrete, in this work, it is assumed that $\mathcal G$ is composed of a grid of $3\times 3$ cells with a bin at each cell center, plus a one-cell wide border.
The position of each pickup station and each bin is known and fixed. 
The warehouse sorts $n_c$ types of parcels ($n_c\leq n_b$), and each type is associated with a set of shipping addresses.
Denote $\mathcal{R}=\{r_1,...,r_n\}$ as the set of robots, $\mathcal{B}=\{b_1,...,b_{n_b}\}$ as the set of bins, $\mathcal{P}=\{p_1,...,p_{n_p}\}$ 
as the set of pickup stations, and $\mathcal{C}=\{1,...,n_c\}$ as the types.
% % 

As a no-load robot arrives at a pickup station, a random package temporarily stored at the station is loaded on the robot.
% 
% There may exist imbalance among the number of packages.
The distribution of parcel types varies for each station; it is assumed that at station $p_k$, parcels of type $j$ arrive with probability $m_{kj}$, $\sum_{j}m_{kj}=1$.
The probability is fixed \emph{a priori} knowledge and can be estimated from historical statistics and continuously updated.
In the bin assignment problem, each bin is associated with a parcel type. 
A \emph{surjection}  $f:\mathcal{B}\rightarrow\mathcal{C}$ must be computed to reduce the average traveling cost. 
There can also be multiple bins for a given type of parcel when $n_b>n_c$.
Proper bin allocation reduces the traveling cost of robots, increasing the throughput.
% 

%In the sorting process, the robot received a task will move to the station, pick up the parcel and deliver it to a correct bin.
% 
As robots continuously pick up and deliver packages, the path-planning process is online and dynamic.
Over the grid-like environment, in each time step, a robot can move up, down, left, and right or wait at its current position.
Collisions must be avoided, specifically: (i) two robots cannot be in the same location at the same time (meet collision), 
(ii) two robots cannot traverse the same edge in opposite directions at the same time (head-on collision), and
(iii) a robot cannot occupy any bin vertex (it will fall into the bin).
% 
%For \dmpp planners, it is required to plan a collision-free path for each robot, routing the robot from its current location to the pickup location and then to the delivery location.
% 
We do not consider the time of loading parcels and dropping parcels.
When a robot finishes a task, it goes to the nearest station to retrieve a new parcel.
The path planning sub-problem in \dmpp is also known as \emph{life-long} multi-agent path finding \cite{stern2019multi}.
The general goal is to let the robots sort the parcels as quickly as possible.
We consider \emph{throughput} as the criteria: the average number of parcels sorted per time step.

\section{Bin Assignment}\label{sec:bin_assignment}
%\TODO{JJ's editing location}
% 
% Also, from a path diversification  point of view, it is better to balance the burdens of each bin as overusing one bin may cause traffic jams and increasing the makespan.
% 
% 
We solve \dmpp in two phases, starting with \emph{bin assignment} that matches each bin with one package type such that each of the $n_c$ types is assigned to at least one bin. If $n_b > n_c$, multiple bins can get the same type. In this phase, we do not consider inter-robot collisions in assigning robots with packages to proper collection bins. 

\textbf{Case 1: $n_b=n_c$}.
Each bin has a unique type; the problem can be unnaturally cast as a \emph{linear assignment} problem \cite{burkard2012assignment}.
The cost of assigning bin $b_i$ to type $j$ denoted as $w_{ij}$ is $$w_{ij}=\sum_{k}m_{kj}dist(p_k,b_i),$$ where  $dist(p_k,b_i)$ is the shortest distance from pickup station $p_k$ to bin $b_i$.
This can be solved optimally by applying the Hungarian algorithm \cite{kuhn1955hungarian} that runs in $O(n_b^3)$ time, sufficiently fast for hundreds of bins.

\textbf{Case 2: $n_b>n_c$}.
In this case, multiple bins may be assigned for taking in the same parcel type, leading to increased sorting throughput and making the problem more complex.
The problem may be modeled as a \emph{generalized assignment problem}, which is NP-hard in general~\cite{ozbakir2010bees}. We propose an optimal (but slow) method and two fast, near-optimal methods for solving the assignment problem. The optimal algorithm is based on mixed integer programming, and sub-optimal ones are greedy and genetic algorithms \cite{golberg1989genetic}. 
\subsection{Optimal Allocation via Mixed Integer Programming}
We propose an optimal mixed integer programming model for bin assignments as: 
% 
%l--> pickup station , j---> type,    i-->bin
%\begin{subequations}
\begin{align}
     \text{Minimize \quad}& 
     \frac{1}{n_p}\sum_{j=1}^{c}\sum_{k=1}^{n_p} m_{kj}d_{kj}\label{eqn:eq_obj}, \text{\quad subject to:}
\end{align}
%\end{subequations}
\begin{subequations}
\begin{align}
     &d_{kj}=\min_{i}\{\frac{1}{x_{ij}}dist(p_k,b_i)\}\label{eqn:eq_dist} \\
         &\sum_{j}x_{ij}=1\quad \text{for each bin $b_i$}\label{eqn:eq_bin}\\
         &\sum_{i}x_{ij}\geq 1\quad \text{for each type $j$}\label{eqn:eq_type}\\
         x_{ij}=&\begin{cases}
         0&\text{if $b_i$ is not used to sort type $j$}\\
         1 &\text{if $b_i$ is used to sort type $j$}
         \end{cases}\label{eqn:eq_binaries}
\end{align}
\vspace{1mm}
\end{subequations}

Eq.~\eqref{eqn:eq_obj} seeks a feasible assignment minimizing the average traveling cost required to sort the parcels, ignoring collisions.
The variable $d_{kj}$ in Eq.~\eqref{eqn:eq_dist} is the minimum traveling cost to sort parcels type $j$, the distance from pickup station $p_k$ to its nearest sorting bin of type $j$.
We let $\frac{1}{x_{ij}}=+\infty$ if  $x_{ij}=0$, i.e., bin $i$ is not reachable as it is not a bin to sort type $j$ parcels.
Eq.~\eqref{eqn:eq_bin}  and Eq.~\eqref{eqn:eq_type} ensures bin-type surjectivity.
We use Gurobi \cite{gurobi} as the solver; we introduce new variables $y_{ij}=\frac{1}{x_{ij}+\varepsilon}dist(p_k,b_i)$, where $\varepsilon$ is a very small positive constant, to avoid divide-by-zero issues.
\subsection{Greedy Bin Allocation}
The above optimal solution scales poorly. Toward speeding up computation, we outline a fast greedy method (Algo.~\ref{alg:greedy}). First, $n_c$ bins are selected by solving a min-cost maximum matching problem (Lines 2-4).
Each assigned bin of type $c$ has cost $w_c$.
Then, we choose bin $b'$ with the highest cost $w_b$, assuming this bin is for sorting parcels of type $c_b$.
We can choose an unassigned bin to share $b'$'s load by iterating over all unassigned bins. For each bin $b''$, we calculate the new traveling cost for sorting type $c$ if adding this bin $b''$ for sorting $c_b$ (Line 9).
Let $\mathbf{B}=\mathbf{b_c}+b''$ be the bins used to sort type $c_b$, the cost is given by 
$\mathcal{C}_b=\sum_{k=1}^{n_p}m_{kc_b}d_{kc_b},$
where
$d_{kc_b}=\min_{b_i\in \mathbf{B}}dist(p_k,b_i).$
The bin with the smallest new traveling cost is assigned for type $c$.
Random assignments are made if the smallest traveling cost exceeds the original cost. The process repeats until all bins are assigned.

\begin{algorithm}
\begin{small}
\DontPrintSemicolon
\SetKwProg{Fn}{Function}{:}{}
\SetKw{Continue}{continue}

 \caption{GreedyAllocation \label{alg:greedy}}
%   \KwIn{robots $A=\{a, $|A| = n = m^2$}
%   \KwIn{Pickup stations $\mathbf{ps}$, bins $\mathbf{b}$, package types $\mathbf{c}$}
  \Fn{\texttt{Greedy}()}{
  \vspace{1mm}
\For{$i\leftarrow[1,...,n_c]$, $j\leftarrow [1,...,n_b]$}{
$\omega_{ij}\leftarrow \sum_{k}m_{kj}dist(p_k,b_j) $\;
}
  \vspace{1mm}
$\{b'\},c_{min}\leftarrow \texttt{minCostMatching}(\omega)$\;
$unallocatedBins\leftarrow\{b\}-\{b'\}$\;
  \vspace{1mm}
\While{$unallocatedBins$ not empty}{
  \vspace{1mm}
$b_{max},c_b\leftarrow$ the bin with largest cost and its type\;
$\mathbf{b_c}\leftarrow$ bins used to sort $c_b$\;
$b''\leftarrow\text{argmin}_{b''}\texttt{ReComputeCost}(c_b,\mathbf{b_c}+b'')$\;
  \vspace{1mm}
\If{newCost $<c_{min}$}{
$c_{min}\leftarrow$newCost\;
assign $b''$ to sort $c_b$\;
remove $b''$ from $unallocatedBins$\;
}
  \vspace{1mm}
\Else{
randomly assign the remaining bins\;
}

}
}

\vspace{1mm}
\end{small}
\end{algorithm}

\subsection{Bin Allocation via Genetic Algorithm}
Genetic algorithms (GA) \cite{holland1992adaptation} can be effective in solving NP-hard optimization problems \cite{golberg1989genetic} and turns out to work nicely on \dmpp. 
% 
%It is an iterative, population-based algorithm and follows the principle of survival of the fittest to search for solutions through the operations of selection, crossover, and mutation. 
% 
In GA, a \emph{chromosome} represents a potential solution, corresponding to bin assignment in our case. 
Specifically, a chromosome is an integer array of length $n_b$. The genome at index $i$ is an integer in $[1,n_c]$ representing the type of the bin $b_i$ is assigned to.
A valid chromosome should contain each type ($n_c$ in total) at least once. 
The \emph{fitness function} is defined as the multiplicative inverse of the average shortest distance of sorting each parcel:
$$F(C)=(\sum_{k=1}^{n_p}\sum_{j=1}^{n_c}m_{kj} d_{kj})^{-1},$$
where $d_{kj}$ is the shortest traveling distance required from pickup station $p_k$ to its nearest bin of type $j$,  
$$d_{kj}=\min_{b_i.type=j}dist(p_k,b_i).$$
The initial \emph{population} of is generated randomly ($80\%$) and by a greedy algorithm ($20\%$).

A new population is generated by \emph{selection} and \emph{crossover} (partially mapped crossover [18]), and \emph{mutation}. At least one best individual is copied without changes to the new population (\emph{elitism selection}), preventing the loss of the current best solution. Rest follows standard GA procedures.

%\begin{remark}
We note that the bin assignment process is \emph{semi-static} in that it remains fixed for a given batch of packages to be sorted but can be updated for each batch. 
%\end{remark}

\section{Distributed Multi-Robot Path Planning}\label{sec:dmrpp}
After bin assignment, robots must be routed to pick up and deliver packages. We present a decentralized method to accomplish this, requiring only that the observable area of a robot is a $3\times 3$ square centered at its current location; it is assumed that a robot can detect potential collisions and communicate with the robots located within this area.
% 
% 
% A $k$-delay conflict $(r_i,r_j,t)$ for robot $r_i$ and $r_j$ is found at timestep $t$ if there exists $\Delta\in [0,k]$  that $v_i(t+\Delta)=v_j(t)$.
% % 
% A $k$-robust plan does not contain any $k$-delay  conflicts \cite{atzmon2018robust}.
% % 
% Considering robustness is important for warehouse application where a plan that can withstand unexpected delays is desirable.
% % 
% In this section we first consider 1-robustness.
% 

\subsection{Directed Road Network Design}
By Robbins' Theorem \cite{robbins1939theorem}, an undirected graph $G(V,E)$ can be oriented to yield a strongly connected directed graph if and only if $G$ is 2-edge-connected, meaning the removal of any single edge from $G$ does not disconnect it. This fundamental insight implies the existence of a strong graph orientation within the context of our warehouse environment.
To facilitate decentralized robot routing, we first impose an orientation on the underlying grid-like environment, transforming it into a strongly connected directed graph that largely preserve shortest paths. 
%To streamline the planning process, we enhance the graph's structure by introducing unidirectional pathways for the warehouse, thereby transforming it into a directed graph.
%In the realm of graph theory, an orientation of an undirected graph involves assigning a direction to each edge, thus converting the original graph into a directed one. The Strong Orientation Problem pertains to the assignment of edge directions in an undirected graph in such a manner that the resulting directed graph maintains strong connectivity.

\begin{figure}[h]
    \centering
    \includegraphics[width=0.8\linewidth]{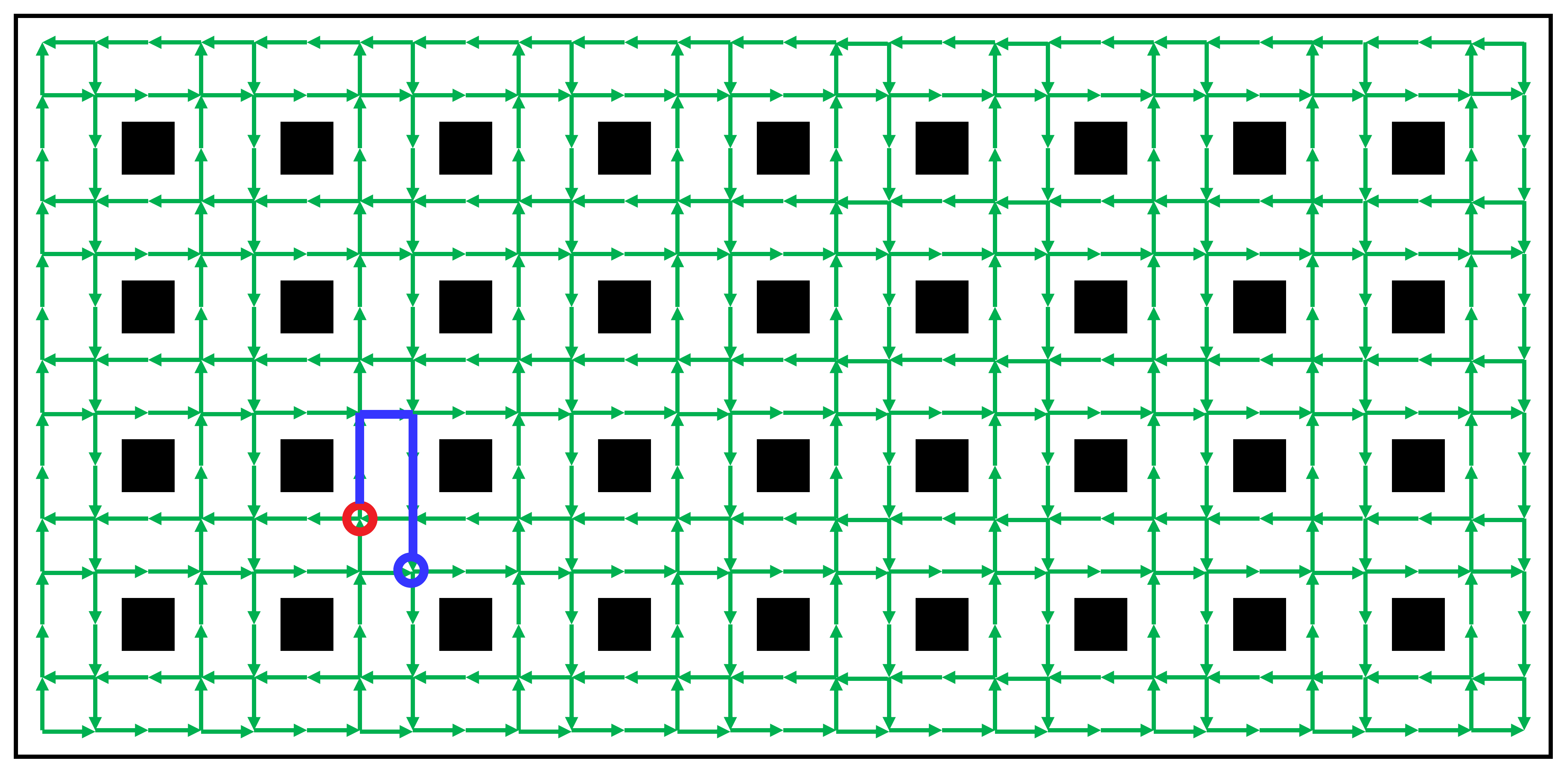}
    \caption{An illustration of the directed network design. To go from the red vertex to the blue on the directed network requires a minimum distance of $7$, incurring an overhead of $5$.}
    \label{fig:directed_graph}
\end{figure}

Employing linear-time depth-first search \cite{Roberts1978}, a strong graph orientation can be identified. However, not all the orientations are good. Determining the optimal strong graph orientation, which minimizes the average pairwise distance, is demonstrated to be NP-hard \cite{plesnik1984sum}. For scenarios resembling city-street grid maps, the investigation into the optimal graph orientation is discussed in \cite{roberts1994optimal}.
We adopt a strategy similar to \cite{roberts1994optimal} to convert the graph into a digraph where the parallel ``streets'' alternate in direction (see Fig. \ref{fig:directed_graph}). 

The directed network provides several crucial advantages. Each edge of the directed network only permits movement in a single direction. Despite solving \mpp optimally is NP-hard on digraphs \cite{nebel2020computational}, the absence of \emph{head-on collisions} greatly reduces computational burden in eliminating considering head-on collisions and reduces the search space's size.
Furthermore, our design incorporating two adjacent embedded "highways" with opposing directions—out of numerous possible configurations does not significantly compromise the optimal solution path. Consider any pair of vertices $u$ and $v$, where $d_d(u,v)$ represents the shortest distance between them in the undirected graph and $d_u(u,v)$ represents the shortest directed distance. Then $d_d(u,v)\leq d_u(u,v)+5$.

\begin{remark}
    Our graph orientation method is not limited to warehouse maps, verifiable using Robbins' Theorem \cite{robbins1939theorem}.
\end{remark}
% 
 %\TODO{will update this figure}

\subsection{Prioritized Recursive Yielding Planner}
We present a decentralized planner over the directed network in which a robot's observation area is limited to a $3\times 3$ square around it.
The robot can detect possible collisions and communicate with other robots in its observation area.
In our case, we only need to address potential meet collisions.
Similar to \ddm \cite{han2020ddm}, our algorithm first uses A* to plan initial paths to goals that may still contain conflicts.
Robots follow the paths and resolve the conflicts locally by letting robots yield to higher-priority robots.

We call the planner \emph{prioritized recursive yielding planner} (\ryp) (Algo.~\ref{alg:ryp}).
\texttt{PlanInitialPaths} generates initial paths using A*.
Robots then execute the paths and resolve conflicts in \texttt{RecursiveMove}.
If a robot reaches its current goal, it receives a new goal and replans the path using A*.
Specifically, if it arrives at a station $p_k$, a new parcel would be randomly generated according to the given distribution $\mathbf{m_k}$.
Then the robot would plan a path to the nearest bin of the corresponding type.
If the robot delivers the parcel, it plans a path to the nearest station to get a new parcel.
\begin{algorithm}
\begin{small}
\DontPrintSemicolon
\SetKwProg{Fn}{Function}{:}{}
\SetKw{Continue}{continue}

 \caption{\ryp \label{alg:ryp}}

 \texttt{PlanInitialPaths()}\;
 \While{True}{
 \For{$i=1\ \text{to}\  n$}{
 \If{$r_i.path$ is empty}{
 \texttt{ReplanPath($r_i$)}\;
 }
 \texttt{RecursiveMove($r_i$)}\;
  \If{$r_i$ does not wait}{$r_i.path.popfront()$\;}
 }
 }
 \end{small}
\end{algorithm}

\begin{figure}[h]
\vspace{-1mm}
    \centering
    \includegraphics[width=.85\linewidth]{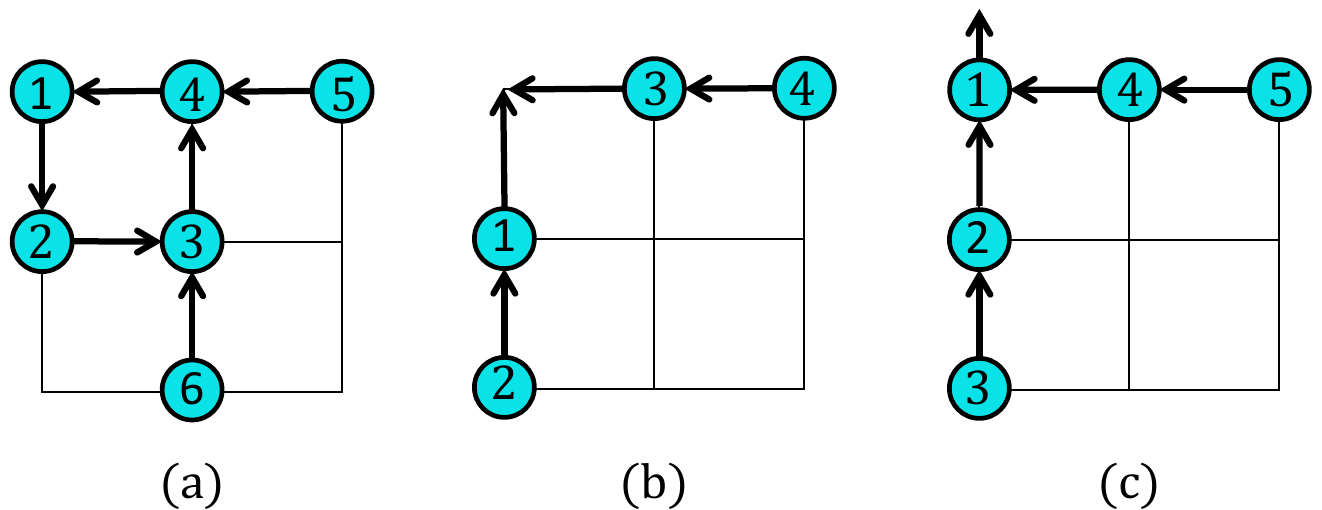}
    \caption{The collision avoiding mechanism. (a) Robots $1$-$4$ are forming a cycle and therefore Robots 5 and 6 should wait. (b)
    Robot 1 tries to move to a free vertex and detects a possible conflict with Robot 3. Robot 3 yields to Robot 1. Robots 1 and robot 2 both move forward while Robots 3 and 4 wait. (c) Robot 1 informs robots 2 and  4 that it can move forward. Robot 4 gives way to Robot 2. Robots 4 and 5 wait at their current position.} 
    \label{fig:planner}
\end{figure}

To resolve potential collisions (\texttt{RecursiveMove}, Algo.~\ref{alg:planner}), first, robots try to figure out whether they form a cycle. To do so, each robot $r_i$ sends a message to its neighbor (if there is one) in the direction it is going, and the message propagates. If robot $r_i$ receives the message back, it is in a cycle.
Those robots that form a cycle have the highest priorities and will move to their next position (see Fig.~\ref{fig:planner}(a)). 

\begin{algorithm}
\begin{small}
\DontPrintSemicolon
\SetKwProg{Fn}{Function}{:}{}
\SetKw{Continue}{continue}

 \caption{\texttt{RecursiveMove($r_i$)} \label{alg:planner}}

%   \KwIn{robots $A=\{a, $|A| = n = m^2$}
  
  \If{$r_i$ is marked}{\Return\;}
  \If{$r_i$ finds it is in a cycle}{
 mark $r_i$ to move to the next vertex and \Return\;
  }
  $r_j\leftarrow$the robot with $\texttt{currentVertex}(r_j)=\texttt{nextVertex}(r_i)$\;
\If{$r_j$!=NULL}{
\texttt{RecursiveMove}($r_j$)\;
\If{$r_j$ is marked to wait}{
mark $r_i$ to wait and \Return\;
}
}
$r_k\leftarrow$ $r_i.\texttt{findConflictedRobot()}$\;
\If{$r_k$= NULL}{mark $r_i$ to move forward and \Return}\;
\If{$r_k$ is marked to move forward}{mark $r_i$ to wait and \Return}\;
Mark one to wait and another to move forward\;
\end{small}
\end{algorithm}
%\vspace{-2mm}

If a robot $r_k$ has no neighbor in the direction it is going, $r_k$ checks with diagonal neighbors to see if there is a meet collision.
For example, in Fig. \ref{fig:planner}(b), robot 1 finds a possible meet collision with robot 3; they communicate to decide who should go first, based on the following policy: If a robot has not delivered its parcel, it has higher priority; otherwise, if a robot has more steps to arrive at its current goal location, it has higher priority. If a tie remains, priority is assigned randomly.
In Fig. \ref{fig:planner}(b), suppose robot 1 has higher priority than robot 3; robot 3 yields to robot 1.
Robot 1 sends a message informing robot 2 it will move to the next location while robot 3 sends a message to tell robot 4 that it has to wait.
For robot $r_k$, if the robot located at its next position is not going to wait at the next step, $r_k$ will communicate with the neighboring conflicted robot and decide who should go first following the above policy (see Fig \ref{fig:planner}(c)).

% \subsection{Properties of the Robust Planner}
% \begin{proposition}
% For a given task, the path returned by the robust planner is pseudo distance-optimal.
% \end{proposition}

% \begin{proposition}
% The robust planner is complete.
% \end{proposition}

% \begin{proposition}
% The time cost of finishing one task for robot $r_i$ never worse than $d_s+n$. 
% \end{proposition}
\subsection{Path Diversification}
In our planner, whenever a meet collision occurs, one of the robots must wait, leading to increased cost.
% 
%This will inevitably increase the traveling cost and decrease the throughput especially when there are many possible conflicts in dense cases.
% 
Therefore, the solution quality is strongly correlated with the number of meet collisions of the planned paths.
Meet collisions can be reduced using a randomized diversification heuristic \cite{han2020ddm}, in which a random path is selected among all candidate paths of the same shortest lengths.
Alternatively, we may let the robot access partial global information, in which case focal search\cite{pearl1982studies} may be applied that uses the number of conflicts as the heuristic to reduce the number of conflicts we need to resolve.
We denote \ryp with path diversification as \emph{enhanced prioritized recursive yielding planner} (\eryp).

\subsection{Probabilistic Deadlocks Prevention Guarantees}
\begin{figure}[h]
\vspace{-3mm}
    \centering
    \includegraphics[width=0.8\linewidth]{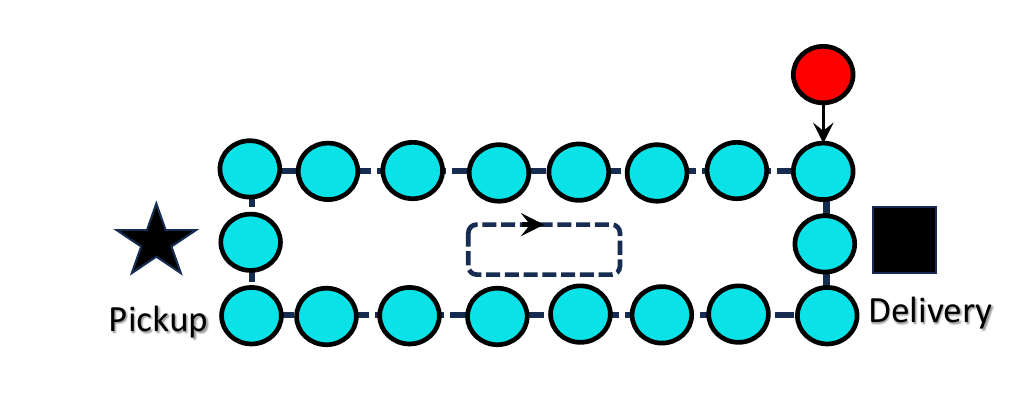}
    \caption{A deadlock may only happen if an infinite cycle of robots is traveling between a pickup station and a matching delivery bin.}
    \label{fig:deadlock_issue}
\end{figure}

Both prioritized planning and decentralized planning may have deadlocks.
A deadlock occurs when a robot cannot reach its designated destination within a finite time.
% 
%This concern becomes particularly relevant in a one-shot framework, where robots remain stationed at their endpoints after completing their tasks.
% 
%In the case of prioritized planning, immobile robots stationed at their objectives can obstruct the movement of other robots, leading to deadlock situations.
% 
%In practical scenarios, robots that complete their tasks typically either vacate the workspace or return to the pickup point to acquire new assignments. This practice ensures that they do not hinder the progress of other robots.
%
When robots are continuously assigned new tasks, as is in our setting, the only potential deadlock arises if a robot attempts to move to a vertex where a cycle full of robots exists (Fig.~\ref{fig:deadlock_issue}).
Robots moving on a cycle have the highest priority; if such cyclic patterns persist indefinitely, robots need to pass vertices on the cycle and cannot progress.
However, infinite cycles can only form if robots move between a fixed pair of a pickup station and the matched delivery bin. This should never happen because these packages at the station do not require sorting. We have the following conclusion. 

\begin{theorem}
\ryp and \eryp algorithms are probabilistically deadlock-free. 
\end{theorem}

% \clearpage
% \section{Best Number Of Robots}
% \subsection{Mean-filed Prediction}
% \subsection{Gradient Descent Method}
\section{Evaluations}\label{sec:evaluation}
We evaluate the proposed bin assignment algorithms and the overall performance of our \dmpp solvers with various \mpp planners, including centralized planner \ecbs($w$=1.5) \cite{barer2014suboptimal}  with horizon cut techniques \cite{li2020lifelong,Han2022OptimizingSU}, \ddm \cite{han2020ddm}, a discrete version of \orca\cite{van2008reciprocal,wang2020mobile}, and the proposed decentralized planners \ryp and \eryp. 
All experiments are performed on an Intel\textsuperscript{\textregistered} Core\textsuperscript{TM} i7-6900K CPU at 3.2GHz. Unless otherwise stated, each data point averages over 20 runs on randomly generated instances.
Assignment algorithms are implemented in Cython, and \mpp planners are implemented in C++.
%A video of four selected simulations can be found at \url{https://youtu.be/F-VLaTWifwo}.
% The source code and evaluation data associated with this research will be made available at \url{https://github.com/arc-l/mrps}, upon the publication of this manuscript. 

% 
\subsection{Comparison of Bin Assignment Algorithms}
\begin{figure}[h]
    \centering
    \includegraphics[width=1.0\linewidth]{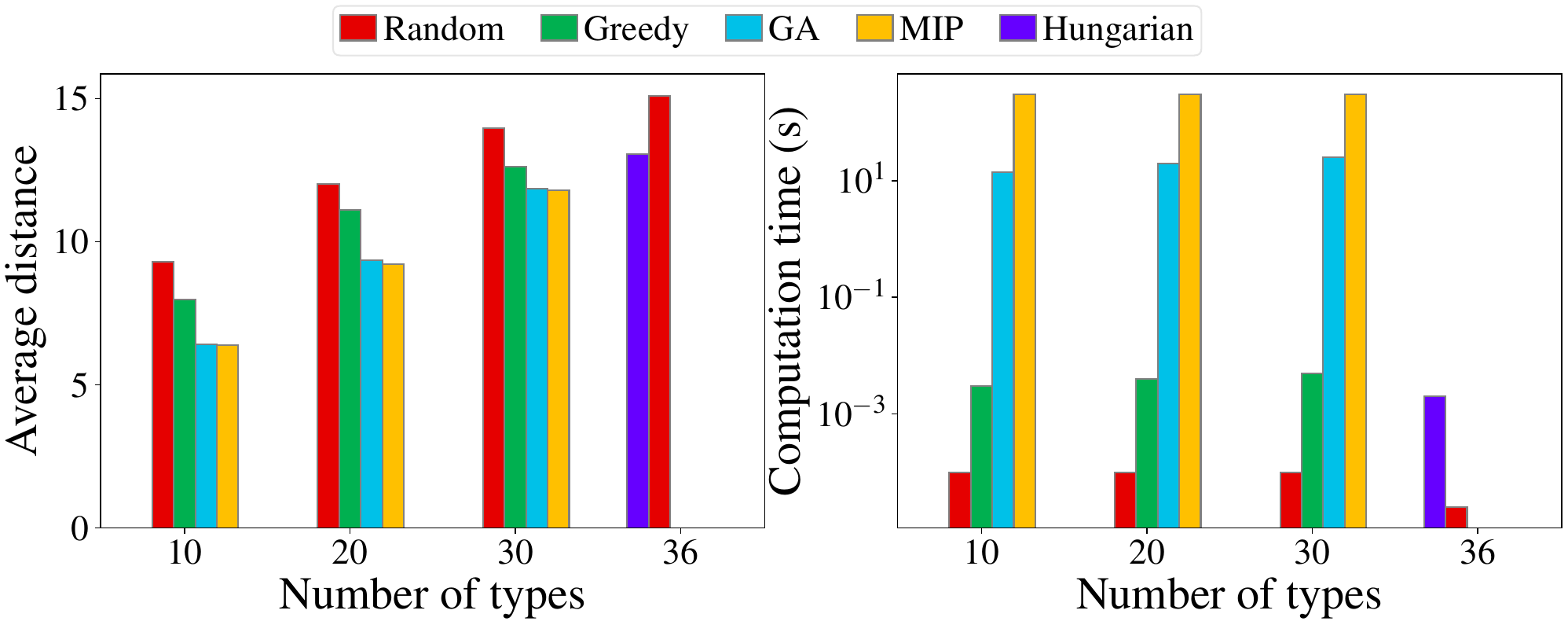}
    \caption{The average distance cost and computation time of different bin assignment algorithms including a randomized, greedy, genetic algorithm (GA), mixed integer programming, and the Hungarian algorithm. For all settings, there are $n_b = 36$ bins and $n_p = 12$ pickup stations. Hungarian algorithm only applies to the case where there are $36$ package types. In this case, we only compare it with the randomized method as it is expected to do better on computation time than the rest of the algorithms.}
    \label{fig:assignment_data}
\end{figure}

We evaluate different assignment algorithms on the sorting warehouse setup shown in Fig. \ref{fig:sorting_center}, which has $n_b = 36$ bins and $n_p = 12$ pickup stations. The number of types $n_c$ varies (but cannot exceed $36$). 
The probability of each type in each station follows a random vector that adds up to $1$. 
We use Gurobi \cite{gurobi} to solve the mixed integer programming (MIP) model with a 5-minute time limit.
For genetic algorithms, the maximum number of iterations is chosen as 800, the population size is set to 100, and the mutation rate is set to 0.08.
The result is presented in Fig. \ref{fig:assignment_data}.
% The result is presented in Table. \ref{table:tb_rth_mpc2}.
% \begin{table}[h!]
% \small\sf\centering
% \centering
% \caption{Average distance and computation time of different assignment algorithms}
% \begin{tabular}{llll}
% \textbf{Average Distance} \\
% \hline
% Number of types&10&20&30\\
% \hline
% Random& 20.4&26.97&30.5\\

% MIP&14.1&20.3&25.4\\

% Greedy&18.2&22.8&26.0\\

% GA&15.0&20.7&25.4\\

% \hline
% \end{tabular}\\[20pt]
% \label{table:tb_rth_mpc2}
% \begin{tabular}{lllll}
% \textbf{Comp Time (s)}\\
% \hline
% Random&MIP&Greedy&GA\\
% \hline
% 0.0002& 300&0.001&21.0\\
% \hline
% \end{tabular}

% \end{table}
Compared to the random assignment, greedy, genetic, and MIP algorithms reduce the average distance by about $10\%$, $25\%$, and $30\%$, respectively.
MIP, while optimal, is slow. Genetic algorithm (GA) runs much faster and achieves nearly identical optimality.
The computation time gap between MIP and GA can be expected to become even bigger as the number of bins increases. 
The result confirms using repetitive bins can reduce the average distance.
When the number of bins equals the number of package types ($n_b = n_c = 36$), the assignment is optimal with the Hungarian algorithm, with faster computation time than greedy, GA, and MIP. 

\subsection{Impact of Assignment Algorithms on \mpp Planners}
\begin{figure}[!htpb]
    \centering
    \includegraphics[width=1.0\linewidth]{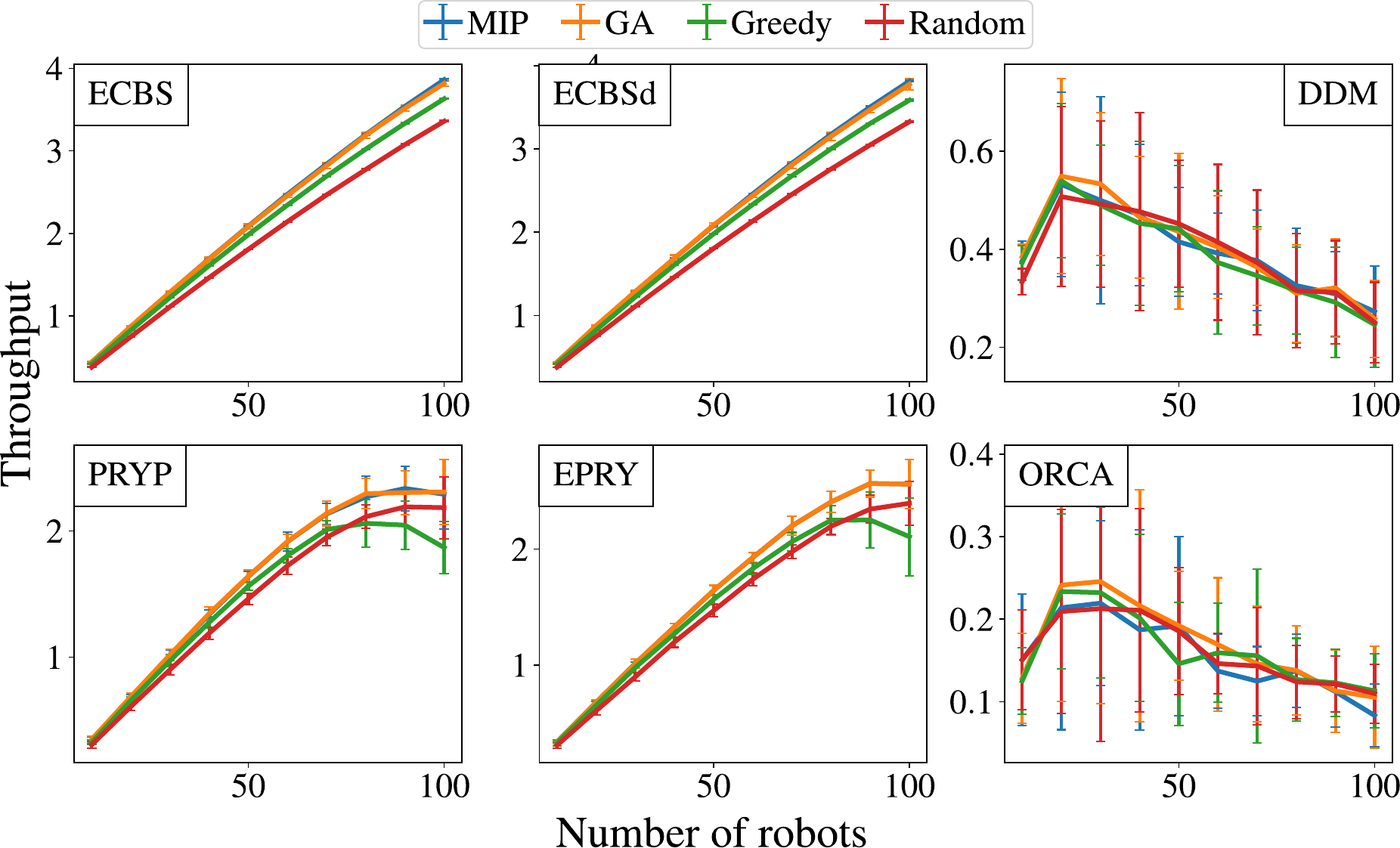}
    \caption{The throughput data of \mpp planners with different bin assignment strategies for different numbers of robots, fixing $n_b = 36, n_p = 12$, and $n_c = 20$. Each figure shows the result for one \mpp algorithm spanning ECBS, decentralized ECBS, DDM, \ryp, \eryp, and \orca.}
    \label{fig:compare_bin}
\end{figure}

%\TODO{Since ECBS and ECBSd yield better throughput, we need to carefully explain that (1) our framework can be combined with any \mpp planner and (2) the overall performance of \ryp can be desirable due to the fast planning time.}
Fixing the number of bins at $n_b = 36$, the number of pickup stations at $n_p = 12$, and the number of package types at $n_c = 20$, and let the number of robots vary, 
we examined the impact of combining different bin assignment algorithms with different \mpp planners. 
The performance of different \mpp planners, \ecbs, \ecbsd, \ddm, \orca, \ryp, \eryp, under different bin assignment strategies are compared. 
\ecbsd is a variant of \ecbs that uses highway heuristics and treats the graph as a directed graph.
For \ecbs and \ecbsd, the planning window and execution window are set to be 5.
The throughput result is shown in Fig.~\ref{fig:compare_bin}. We omit the computation time comparison, which is not important. 

We observe that the GA, MIP, and greedy approaches all improve the throughput of \ecbs and \ecbsd about $10\%-20\%$ compared to random assignment. GA and MIP also clearly outperform greedy assignments. 
Note that the throughput equals $n/\Bar{d}$, where $\Bar{d}$ is the average traveling cost required to deliver and sort a parcel.
As \ecbs and \ecbsd are bounded suboptimal, the solution costs are close to the traveling distance. Combined with the bin assignment algorithms, the throughput of \ecbs and \ecbsd scale nearly linearly to the number of robots within the range (but we will see that they are more computationally demanding when compared to decentralized planning methods).

\ddm, \ryp, and \eryp resolve conflicts locally, leading to improved throughput  at lower robot density.
At higher robot density, due to the high number of conflicts to be resolved, $\Bar{d}$ would increase too with respect to the number of robots.
Consequently, these algorithms have a peak throughput, indicating the number of robots that can achieve the best throughput.
Compared to \ddm, \ryp and \eryp have much better throughput.
% 

% \TODO{We need to have several environment sizes - at least 4, with 1 smaller and two larger.}

% \TODO{We probably should add comparison to ORCA, since DDM is not very suitable for this type of environment.}

\subsection{Comparisons of \dmpp Solvers on Larger Maps}
\begin{figure}[h]
    \centering
    \includegraphics[width=.95\linewidth]{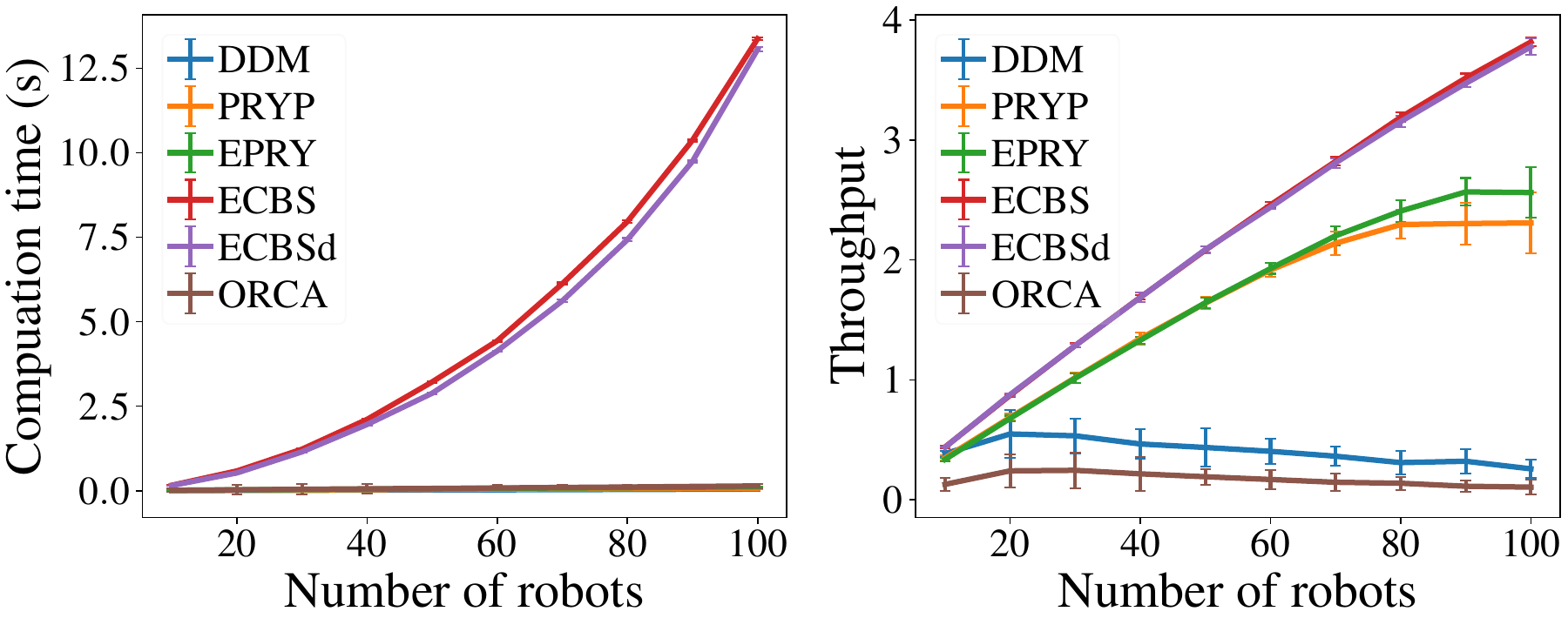}
    \caption{Computation time and throughput of \dmpp solvers using GA for bin assignment on a $14\times 29$ map with $n_p = 12$ stations and $n_b = 36$ bins for sorting $n_c = 20$ types of parcels. The \mpp solvers used include \ecbs, \ecbsd, \ddm, \orca, \ryp, and \eryp. It is straightforward to observe that \ryp and \eryp compute solutions extremely fast and at the same time realize throughput that is fairly close to ECBS.}
    \label{fig:compareDmpp}
\end{figure}

\begin{figure}[h]
    \centering
    \includegraphics[width=0.95\linewidth]{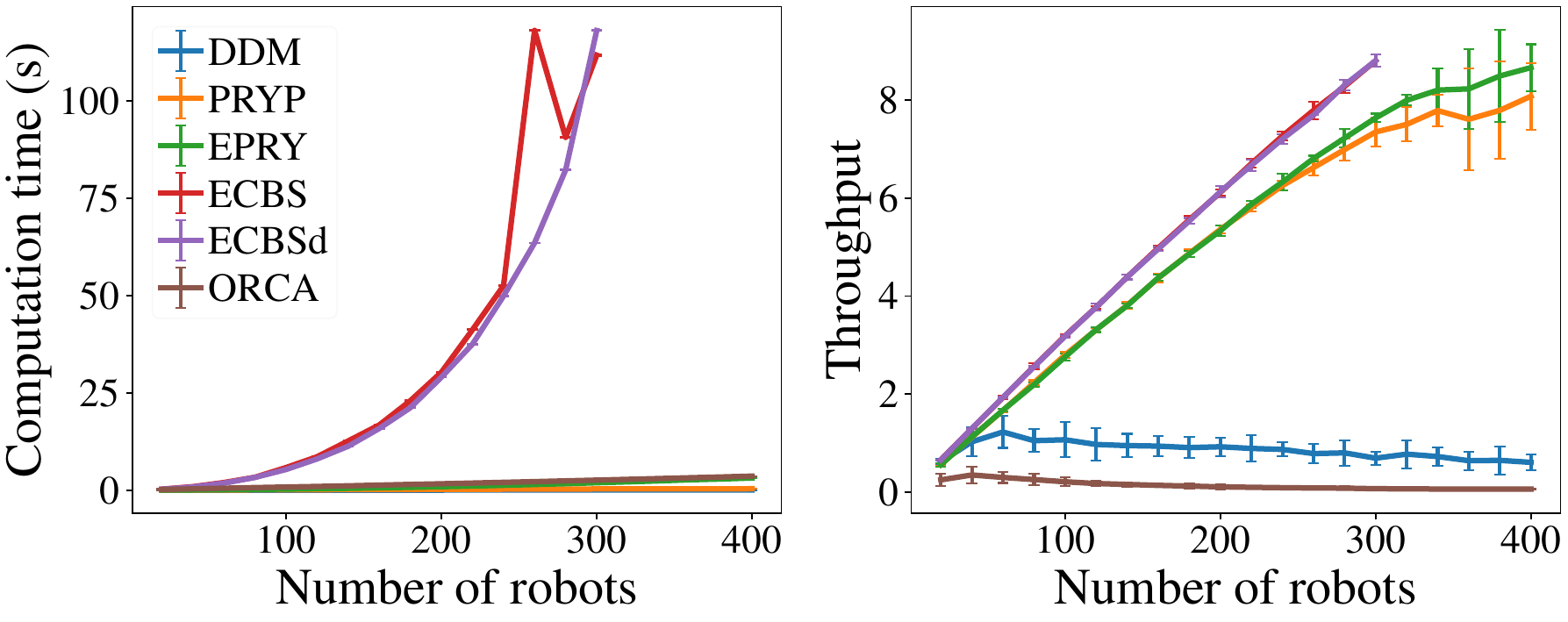}
    \caption{Computation time and throughput of \dmpp solvers using GA for bin assignment on a $32\times 62$ map with $n_p = 20$ stations and $n_b = 300$ bins for sorting $n_c = 100$ types of parcels. Up to $400$ robots are tested. The \mpp algorithms used are again ECBS, decentralized ECBS, DDM, \ryp, \eryp, and \orca. As the number of robots exceeds $300$, ECBS algorithms take too much time to compute a solution.}
    \label{fig:compareDmppMedium}
\end{figure}

\begin{figure}[h]
    \centering
    \includegraphics[width=0.95\linewidth]{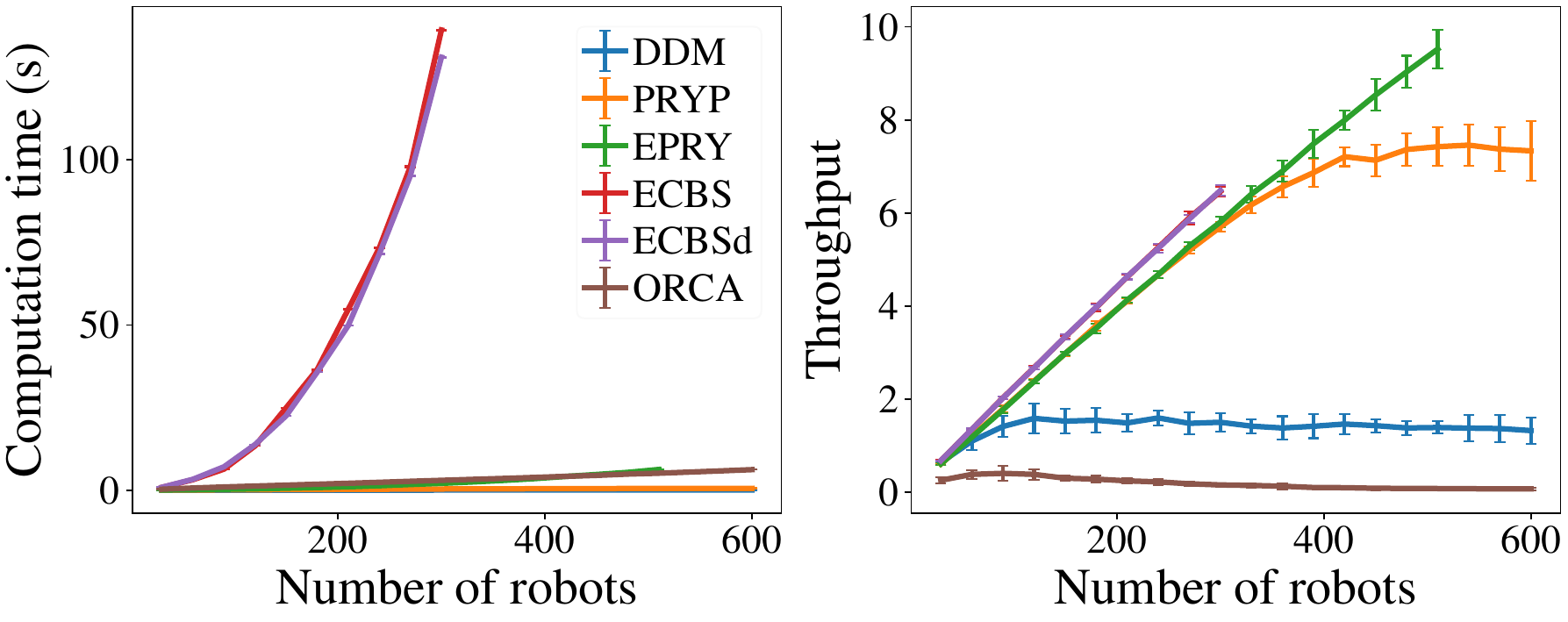}
    \caption{Computation time and throughput of \dmpp solvers using GA for bin assignment on a large $47\times 92$ map with $n_p = 30$ stations and $n_b = 450$ bins for sorting $n_c = 200$ types of parcels. This evaluation largely agrees with Fig.~\ref{fig:compareDmppMedium} but suggests that \eryp outperforms \ryp as the problem becomes even harder.}
    \label{fig:compareDmppLarge}
\end{figure}

Lastly, we present evaluation results (Fig.~\ref{fig:compareDmpp}-Fig.~\ref{fig:compareDmppLarge}) on full \dmpp solvers using GA for bin assignment. Specifically, we evaluate scalability, computation time, and the achieved throughput. 
For evaluating scalability, three maps are used. 
We let each \dmpp planner run 500 steps for each setting and calculate the computation time and average throughput.
% using the same setting as the previous section.
% 
%The results are shown in Fig.\ref{fig:compareDmpp}-\ref{fig:compareDmppLarge}.

For the smallest map (Fig.~\ref{fig:compareDmpp}), \ecbs and \ecbsd have the best throughput since they plan the paths and resolve the conflicts centrally.
% 
%\ecbsd does not need to resolve head-on conflicts because of  using the highway heuristics.
% 
%Therefore, \ecbsd is slightly faster than \ecbs and adopts the almost the same throughput.
% 
%However, due to the NP-hardness of  \dmpp, they are not very scalable with respect to the number of robots.
On the other hand, these methods take the most time, rendering them impractical as the map size and the number of robots increase (Fig.~\ref{fig:compareDmppMedium} and Fig.~\ref{fig:compareDmppLarge}). For example, for the largest map (Fig.~\ref{fig:compareDmppLarge}) with $300$ robots, it takes over $0.5$ second to compute a solution for routing each robot, which is prohibitively expensive for online applications.  % 

\ddm, \ryp, \eryp, and \orca are much more scalable, making them more suitable for online applications such as parcel sorting.
% 
%This is at the cost of global optimality (throughput), however.
% 
While \ddm and \orca have fairly sub-optimal performance in terms of throughput, \ryp and \eryp deliver throughputs that are directly comparable to centralized methods (while the centralizes are still scalable).
% 
% 
%Note that \ryp can be decentralized without the global information.
% 
\eryp adopts the information of the planned paths of other robots and applies focal search to reduce the number of conflicts to resolve; therefore, \eryp achieves better throughput in dense cases as compared with \ryp.
All in all, \ryp and \eryp are the most suitable for solving \dmpp tasks.

%%%%%%%%%%%%%%%%%%%%%%%%%%%%%%%%%%%%%%%%%%%%%%%%%%%%%%%%%%%%%%%%%%%%%%%%%%%%%%%%

%%%%%%%%%%%%%%%%%%%%%%%%%%%%%%%%%%%%%%%%%%%%%%%%%%%%%%%%%%%%%%%%%%%%%%%%%%%%%%

\section{Conclusion}\label{sec:conclusion}
In this research, we tackle multi-robot parcel sorting (\dmpp), partitioning it into two phases: bin assignment and multi-robot routing.
We propose several effective algorithms for bin assignment that significantly reduce the average traveling distance, leading to increased throughput. These algorithms can be combined with any multi-robot path planning routines.  
For the multi-robot routing phase of \dmpp, we propose a prioritized, probabilistically deadlock-free algorithm over a directed network. The decentralized approach, integrated into an \dmpp solver,  achieves excellent overall performance in terms of throughput and scalability as compared to other advanced \dmpp solvers.
% 
%Last but not least, it's worth noting that our assignment and planning algorithms are not restricted to sorting warehouse but can also be applied in other types of grid-like settings, i,e, fulfillment warehouses. 

%In future work, we will consider robustness \cite{atzmon2018robust} and rotations in the planning process. 
% 
%In practice, following the original plan may not be possible due to unexpected events that delay some of the robots.
% 
%Therefore, a robust plan that can be followed even if unpredictable delays occur is essential.
% 
%Also, we would like to exploit learning-based approaches to study the behavior of robots and improve the coordination policy in a decentralized manner.



% \section*{ACKNOWLEDGMENT}

% \bibliographystyle{plainnat}
\bibliographystyle{IEEEtran}
% %\bibliography{references}
\bibliography{all}
%%%%%%%%%%%%%%%%%%%%%%%%%%%%%%%%%%%%%%%%%%%%%%%%%%%%%%%%%%%%%%%%%%%%%%%%%%%%%%%%

\end{document}